\documentclass[11pt,a4paper]{article}
\usepackage[hyperref]{emnlp2018}
\usepackage{times}
\usepackage{latexsym}

\usepackage{url}

\aclfinalcopy 

\usepackage{booktabs}

\usepackage{graphicx}  
\usepackage{amsmath} 
\usepackage{amsfonts} 
\usepackage{bm} 
\usepackage{multirow} 
\def\argmax{\mathop{\rm argmax}}%

\title{A Stable and Effective Learning Strategy for Trainable Greedy Decoding}

\author{Yun Chen$^\dagger$, Victor O.K. Li$^\dagger$, Kyunghyun Cho$^{\ddagger,\star}$, Samuel R. Bowman$^\ddagger$  \\ 
$^\dagger$The University of Hong Kong, $^\ddagger$New York University,
$^\star$CIFAR Global Scholar \\
\texttt{yun.chencreek@gmail.com}, \texttt{vli@eee.hku.hk},\\
\texttt{kyunghyun.cho@nyu.edu}, \texttt{bowman@nyu.edu}\\} 
\date{}

\begin{document}
\maketitle
\begin{abstract}
Beam search is a widely used approximate search strategy for neural network decoders, and it generally outperforms simple greedy decoding on tasks like machine translation. However, this improvement comes at substantial computational cost.
In this paper, we propose a flexible new method that allows us to reap nearly the full benefits of beam search with nearly no additional computational cost. The method revolves around a small neural network \textit{actor} that is trained to observe and manipulate the hidden state of a previously-trained decoder.
To train this actor network, we introduce the use of a pseudo-parallel corpus built using the output of beam search on a base model, ranked by a target quality metric like BLEU. Our method is inspired by earlier work on this problem, but requires no reinforcement learning, and can be trained reliably on a range of models. Experiments on three parallel corpora and three architectures show that the method yields substantial improvements in translation quality and speed over each base system.
\end{abstract}

\section{Introduction}

Neural network sequence decoders yield state-of-the-art results for many text generation tasks, including machine translation \citep{Bahdanau2014NeuralMT, Luong2015EffectiveAT, Gehring2017ConvolutionalST, Vaswani2017AttentionIA, mostafa2018}, text summarization \citep{rush2015neural, Ranzato2015SequenceLT, see-liu-manning:2017:Long, paulus2017deep} and image captioning \citep{vinyals2015show, xu2015show}. These decoders generate tokens from left to right, at each step giving a distribution over possible next tokens, conditioned on both the input and all the tokens generated so far. However, since the space of all possible output sequences is infinite and grows exponentially with sequence length, heuristic search methods such as greedy decoding or beam search \citep{Graves2012SequenceTW, BoulangerLewandowski2013AudioCR} must be used at decoding time to select high-probability output sequences. Unlike greedy decoding, which selects the token of the highest probability at each step, beam search expands all possible next tokens at each step, and maintains the $k$ most likely prefixes, where $k$ is the beam size. Greedy decoding is very fast---requiring only a single run of the underlying decoder---while beam search requires an equivalent of $k$ such runs, as well as substantial additional overhead for data management. However, beam search often leads to substantial improvement over greedy decoding. For example, \citet{Ranzato2015SequenceLT} report that beam search (with $k=10$) gives a 2.2 BLEU improvement in translation and a 3.5 ROUGE-2 improvement in summarization over greedy decoding.

Various approaches have been explored recently to improve beam search by improving the method by which candidate sequences are scored \cite{li2016simple, shu2017later}, the termination criterion \cite{huang2017finish}, or the search function itself \cite{Li2017LearningTD}. In contrast, \citet{Gu2017TrainableGD} have tried to directly improve greedy decoding to decode for an arbitrary decoding objective. They add a small \textit{actor} network to the decoder and train it with a version of policy gradient to optimize sequence objectives like BLEU. However, they report that they are seriously limited by the instability of this approach to training.

In this paper, we propose a procedure to modify a trained decoder to allow it to generate text greedily with the level of quality (according to metrics like BLEU) that would otherwise require the relatively expensive use of beam search. To do so, we follow \citet{cho2016noisy} and \citet{Gu2017TrainableGD} in our use of an \textit{actor} network which manipulates the decoder's hidden state, but introduce a stable and effective procedure to train this actor. In our training procedure, the actor is trained with ordinary backpropagation on a model-specific artificial parallel corpus. This corpus is generated by running the un-augmented model on the training set with large-beam beam search, and selecting outputs from the resulting $k$-best list which score highly on our target metric. 

Our method can be trained quickly and reliably, is effective, and can be straightforwardly employed with a variety of decoders. We demonstrate this for neural machine translation on three state-of-the-art architectures: RNN-based \citep{Luong2015EffectiveAT}, ConvS2S \citep{Gehring2017ConvolutionalST} and Transformer \citep{Vaswani2017AttentionIA}, and three corpora: IWSLT16 German-English,\footnote{https://wit3.fbk.eu/} WMT15 Finnish-English\footnote{http://www.statmt.org/wmt15/translation-task.html} and WMT14 German-English.\footnote{ http://www.statmt.org/wmt14/translation-task}

\section{Background}
\subsection{Neural Machine Translation}
In sequence-to-sequence learning, we are given a set of source--target sentence pairs and tasked with learning to generate each target sentence (as a sequence of words or word-parts) from its source sentence. We first use an encoding model such as a recurrent neural network to transform a source sequence into an encoded representation, then generates the target sequence using a neural decoder. 

Given a source sentence $\mathbf{x}=\{x_1,...,x_{T_s}\}$, a neural machine translation system models the distribution over possible output sentences $\mathbf{y}=\{y_1,...,y_{T}\}$ as:
\begin{eqnarray}
P(\mathbf{y}|\mathbf{x}; \bm{\theta})=\prod_{t=1}^T P(y_t|y_{<t},\mathbf{x}; \bm{\theta}), 
\end{eqnarray}
where $\bm{\theta}$ is the set of model parameters.

Given a parallel corpus $D_{x,y}$ of source--target sentence pairs, the neural machine translation model can be trained by maximizing the log-likelihood:
\begin{eqnarray}
\hat{\bm{\theta}} = \argmax_{\bm{\theta}} \Bigg\{  \sum_{\langle \mathbf{x}, \mathbf{y} \rangle \in D_{x,y} } \log P(\mathbf{y}|\mathbf{x}; \bm{\theta}) \Bigg\}. 
\end{eqnarray}

\subsection{Decoding} 
\begin{figure*}[!t]
	\centering\includegraphics[width=0.9\linewidth]{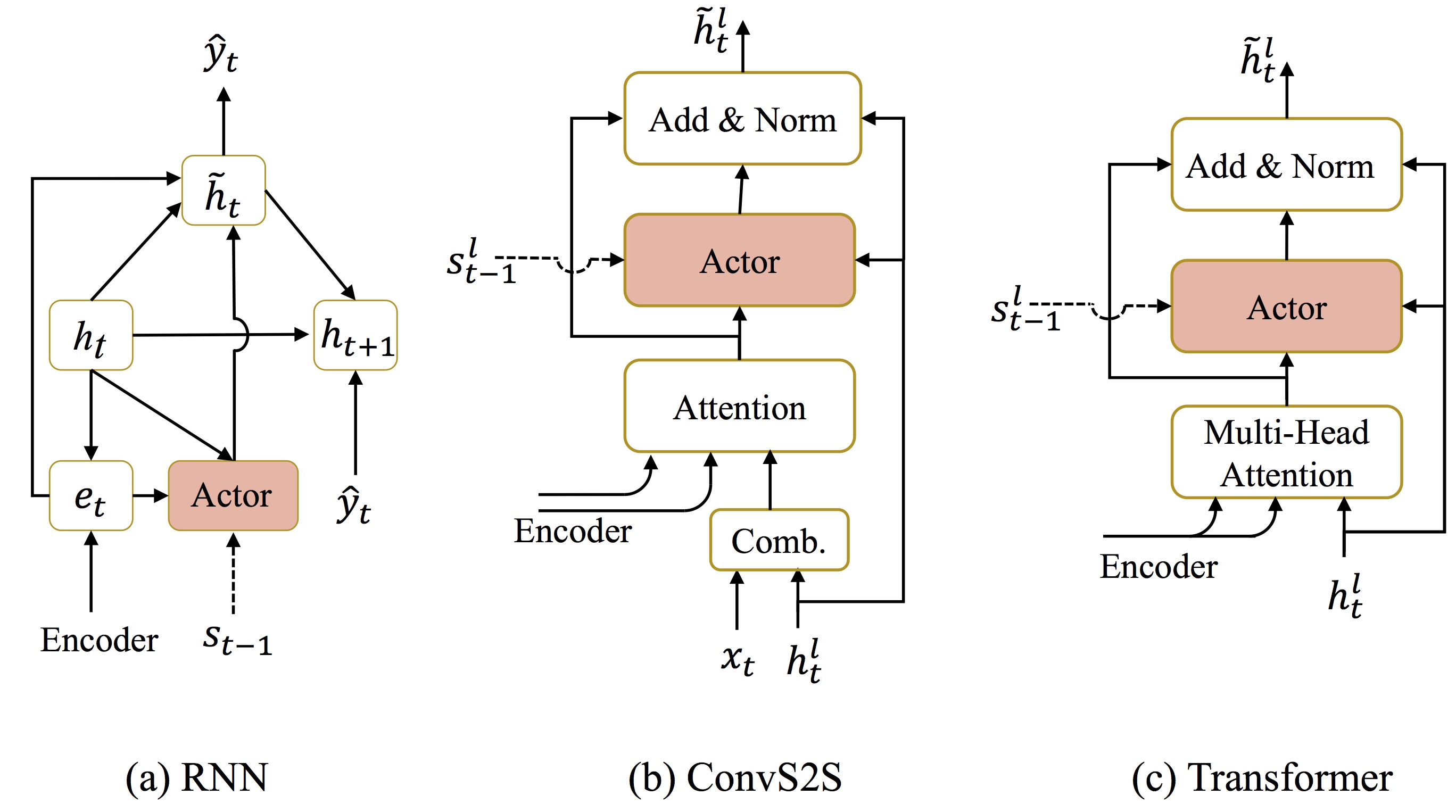}
	\caption{A single step of a generic actor interacting with a decoder of each of three types. The dashed arrows denote an optional recurrent connection in the actor network.}\label{fig:sys}
\end{figure*} 

Given estimated model parameters $\hat{\bm{\theta}}$, the decision rule for finding the translation with the highest probability for a source sentence $\mathbf{x}$ is given by
\begin{eqnarray}
\hat{\mathbf{y}} =\argmax_{\mathbf{y}}\left\{ P(\mathbf{y}|\mathbf{x}; \hat{\bm{\theta}}) \right\} . \label{std:decode}
\end{eqnarray}
However, since such exact inference requires the intractable enumeration of large and potentially infinite set of candidate sequences, we resort to approximate decoding algorithms such as greedy decoding, beam search, noisy parallel decoding \citep[NPAD;][]{cho2016noisy}, or trainable greedy decoding \citep{Gu2017TrainableGD}.

\paragraph{Greedy Decoding} In this algorithm, we generate a single sequence from left to right, by choosing the token that is most likely at each step. The output $\hat{\mathbf{y}}=\{\hat{y}_1,...,\hat{y}_{T}\}$ can be represented as
\begin{eqnarray}
\hat{y}_t=\argmax_{y} \left\{ P(y|\hat{y}_{<t},\mathbf{x};\hat{\bm{\theta}})\right\}.
\end{eqnarray}
Despite its low computational complexity of $O(|V| \times T)$, the translations selected by this method may be far from optimal under the overall distribution given by the model.

\paragraph{Beam Search}
Beam search decodes from left to right, and maintains $k>1$ hypotheses at each step. At each step $t$, beam search considers all possible next tokens conditioned on the current hypotheses, and picks the $k$ with the overall highest scores $\prod_{t'=1}^t P(y_{t'} | y_{<t'}, \mathbf{x};\hat{\bm{\theta}})$. When all the hypotheses are complete (they end in an end-of-the-sentence symbol or reach a predetermined length limit), it returns the hypothesis with the highest likelihood. Tuning to find a roughly optimal beam size $k$ can yield improvements in performance with sizes as high as 30 \cite{koehn-knowles:2017:NMT, britz-EtAl:2017:EMNLP2017}. However, the complexity of beam search grows linearly in beam size, with high constant terms, making it undesirable in some applications where latency is important, such as in on-device real-time translation.

\paragraph{NPAD}
Noisy parallel approximate decoding \citep[NPAD;][]{cho2016noisy} is a parallel decoding algorithm that can be used to improve greedy decoding or beam search. The main idea is that a better translation with a higher probability may be found by injecting unstructured random noise into the hidden state of the decoder network. Positive results with NPAD suggest that small manipulations to the decoder hidden state can correspond to substantial but still reasonable changes to the output sequence.

\paragraph{Trainable Greedy Decoding}
Approximate decoding algorithms generally approximate the maximum-a-posteriori inference described in Equation \ref{std:decode}. This is not necessarily the optimal basis on which to generate text, since (i) the conditional log-probability assigned by a trained NMT model does not necessarily correspond well to translation quality \cite{Tu2017NeuralMT}, and (ii) different application scenarios may demand different decoding objectives \cite{Gu2017TrainableGD}. To solve this, \citet{Gu2017TrainableGD} extend NPAD by replacing the unstructured noise with a small feedforward actor neural network. This network is trained using a variant of policy gradient reinforcement learning to optimize for a target quality metric like BLEU under greedy decoding, and is then used to guide greedy decoding at test time by modifying the decoder's hidden states. Despite showing gains over the equivalent actorless model, their attempt to directly optimize the quality metric makes training unstable, and makes the model nearly impossible to optimize fully. This paper offers a stable and effective alternative approach to training such an actor, and further develops the architecture of the actor network.

\section{Methods}

We propose a method for training a small actor neural network, following the trainable greedy decoding approach of \citet{Gu2017TrainableGD}. This actor takes as input the current decoder state $\mathbf{h}_t$, an attentional context vector $\mathbf{e}_t$ for the source sentence, and optionally the previous hidden state $\mathbf{s}_{t-1}$ of the actor, and produces a vector-valued action $\mathbf{a}_t$ which is used to update the decoder hidden state. The actor function can take on a variety of forms, and we explore four: a feedforward network with one hidden layer (\textit{ff}), feedforward network with two hidden layers (\textit{ff2}), a GRU recurrent network \citep[\textit{rnn};][]{cho-EtAl:2014:EMNLP2014}, and gated feedforward network (\textit{gate}). 

The feedforward \textit{ff} actor function is computed as
\begin{eqnarray}
\mathbf{z}_t &=&  \sigma ([\mathbf{h}_t,\mathbf{e}_t] W^i+\mathbf{b}^i), \nonumber \\
\mathbf{a}_t &=& \tanh (\mathbf{z}_t W^o+\mathbf{b}^o), 
\label{eqn:ff}
\end{eqnarray}
the \textit{ff2} actor is computed as 
\begin{eqnarray}
\mathbf{z}_t^1 &=&  \sigma ([\mathbf{h}_t,\mathbf{e}_t] W^i+\mathbf{b}^i), \nonumber \\
\mathbf{z}_t^2 &=&  \sigma (\mathbf{z}_t^1 W^z+\mathbf{b}^z), \nonumber \\
\mathbf{a}_t &=& \tanh (\mathbf{z}_t^2 W^o+\mathbf{b}^o),
\label{eqn:ff2}
\end{eqnarray}
the \textit{rnn} actor is computed as
\begin{eqnarray}
 \mathbf{z}_t&=& \sigma ([\mathbf{h}_t,\mathbf{e}_t] U^z + \mathbf{s}_{t-1} W^z ), \nonumber \\
 \mathbf{r}_t&=& \sigma ([\mathbf{h}_t,\mathbf{e}_t] U^r +\mathbf{s}_{t-1} W^r ), \nonumber \\
 \mathbf{\tilde{s}}_t&=& \tanh \big([\mathbf{h}_t,\mathbf{e}_t] U^h + (\mathbf{s}_{t-1} \circ \mathbf{r}_t) W^h \big), \nonumber \\
\mathbf{s}_t&=& (1-\mathbf{z}_t) \circ \mathbf{\tilde{s}}_t + \mathbf{z}_t \circ \mathbf{s}_{t-1}, \nonumber \\
\mathbf{a}_t &=& \mathbf{s}_t U,
 \label{eqn:rnn}
\end{eqnarray}
and the \textit{gate} actor is computed as 
\begin{eqnarray}
\mathbf{z}_t &=&  \sigma ([\mathbf{h}_t,\mathbf{e}_t] U^z), \nonumber \\
\mathbf{a}_t &=& \mathbf{z}_t \circ \tanh ([\mathbf{h}_t,\mathbf{e}_t] U). 
\label{eqn:gate}
\end{eqnarray}

Once the action $\mathbf{a}_t$ has been computed, the hidden state $\mathbf{h}_t$ is simply replaced with the updated state $\mathbf{\tilde{h}}_t$:
\begin{equation}
\mathbf{\tilde{h}}_t =  f(\mathbf{h}_t,\mathbf{e}_t)+\mathbf{a}_t.
\end{equation}

Figure~\ref{fig:sys} shows a single step of the actor interacting with the underlying neural decoder of each of the three NMT architectures we use: the RNN-based model of \citet{Luong2015EffectiveAT}, ConvS2S \citep{Gehring2017ConvolutionalST}, and Transformer \citep{Vaswani2017AttentionIA}. We add the actor at the decoder layer immediately after the computation of the attentional context vector. For the RNN-based NMT, we add the actor network only to the last decoder layer, the only place attention is used. Here, it takes as input the hidden state of the last decoder layer $\mathbf{h}_t^L$ and the source context vector $\mathbf{e}_t$, and outputs the action $\mathbf{a}_t$, which is added back to the attention vector $\tilde{\mathbf{h}}_t^L$. For ConvS2S and Transformer, we add an actor network to each decoder layer. This actor is added to the sublayer which performs multi-head or multi-step attention over the output of the encoder stack. It takes as input the decoder state $\mathbf{h}_t^l$ and the source context vector $\mathbf{e}_t^l$, and outputs an action $\mathbf{a}_t^l$ which is added back to get $\tilde{\mathbf{h}}_t^l$.  

\paragraph{Training}
To overcome the severe instability reported by \citet{Gu2017TrainableGD}, we introduce the use of a \textit{pseudo-parallel corpus} generated from the underlying NMT model \citep{gao-he:2013:NAACL-HLT, auli-gao:2014:P14-2, kim-rush:2016:EMNLP2016,P17-1176,Freitag2017EnsembleDF,Zhang2017TowardsCA} for actor training. This corpus includes pairs that both (i) have a high model likelihood, so that we can coerce the model to generate them without much additional training or many new parameters and, (ii) represent high-quality translations, measured according to a target metric like BLEU. We do this by generating sentences from the original unaugmented model with large-beam beam search and selecting the best sentence from the resulting $k$-best list according to the decoding objective.

More specifically, let $\langle \mathbf{x}, \mathbf{y} \rangle$ be a sentence pair in the training data and $\mathcal{Z}=\{\mathbf{z}^1,...,\mathbf{z}^k\}$ be the $k$-best list from beam search on the pretrained NMT model, where $k$ is the beam size. We define the objective score of the translation $\mathbf{z}$ w.r.t. the gold-standard translation $\mathbf{y}$ according to a target metric such as BLEU \cite{Papineni2002BleuAM}, NIST \cite{doddington2002automatic}, negative TER \cite{snover2006study}, or METEOR \cite{lavie2009meteor} as $O(\mathbf{z}, \mathbf{y})$. Then we choose the sentence $\tilde{\mathbf{z}}$ that has the highest score to become our new target sentence:
\begin{equation}
\tilde{\mathbf{z}} =  \argmax_{\mathbf{z}=\mathbf{z}^1,..,\mathbf{z}^k} {O(\mathbf{z}, \mathbf{y}}).
\end{equation}

Once we obtain the pseudo-corpus $D_{x,z}=\{\langle \mathbf{x_i}, \tilde{\mathbf{z}}_i\}_{i=1}^n \}$, we keep the underlying model fixed and train the actor by maximizing the log-likelihood of the actor parameters with these pairs:
\begin{eqnarray}
\bm{\hat{\theta}}_{a} = 
\argmax_{\bm{\theta}_{a}}  
  \Bigg\{\sum_{\langle \mathbf{x}, \mathbf{z} \rangle \in D_{x,z} } \log P(\mathbf{z}|\mathbf{x}; \bm{\hat{\theta}},\bm{\theta}_{a}) \Bigg\} \nonumber \\
   \qquad \qquad \qquad \qquad \qquad \qquad \qquad 
\end{eqnarray} 
In this way, the actor network is trained to manipulate the neural decoder's hidden state at decoding time to induce it to produce better-scoring outputs under greedy or small-beam decoding.

\section{Experiments}
\begin{table*}[t]
\small
\begin{center}
\renewcommand\tabcolsep{5.0pt}
\begin{tabular}{l||ccc|ccc||ccc|ccc}
& \multicolumn{3}{c}{BLEU$\uparrow$} & \multicolumn{3}{c|}{tok/s$\uparrow$} & \multicolumn{3}{c}{BLEU$\uparrow$} & \multicolumn{3}{c}{tok/s$\uparrow$} \\ 
& \multicolumn{1}{c}{greedy} & \multicolumn{1}{c}{beam4} & \multicolumn{1}{c}{tg} &  \multicolumn{1}{c}{greedy} & \multicolumn{1}{c}{beam4} & \multicolumn{1}{c|}{tg} & \multicolumn{1}{c}{greedy} & \multicolumn{1}{c}{beam4} & \multicolumn{1}{c}{tg} & \multicolumn{1}{c}{greedy} & \multicolumn{1}{c}{beam4} & \multicolumn{1}{c}{tg}
\\ 
\toprule
IWSLT16 & \multicolumn{6}{c|}{ De $\rightarrow$ En}  & \multicolumn{6}{c}{En $\rightarrow$ De} \\ 
\midrule
RNN  & 23.57  & \underline{24.90} & 23.59 & 62.8 & 45.0 & 60.4 & 20.05  & \underline{21.11} & 19.88 & 48.1 & 32.5 & 45.7    \\
ConvS2S  & 27.44  & \underline{28.80} & 28.74 & 191.1 & 87.2 & 167.5 & 22.88  & 24.02 & \underline{24.42} & 136.5 & 64.0 & 124.0   \\
Transformer & 27.15  & \underline{28.74}  & 28.36 &  63.9 & 31.0 & 59.8 & 23.87  & 25.03 & \underline{25.46} & 57.9 & 26.5 & 51.2 \\ 
\toprule
WMT15 & \multicolumn{6}{c|}{ Fi $\rightarrow$ En}  & \multicolumn{6}{c}{En $\rightarrow$ Fi} \\ 
\midrule
RNN  & 12.45  & \underline{13.22} & 13.02 & 51.5 & 33.1 & 43.4 & 9.77  & \underline{10.81} & 10.57 & 44.0 & 31.2 & 43.8    \\
ConvS2S  & 15.43  & 16.86 & \underline{17.17} & 24.8 & 11.4 & 16.2 & 12.65  & 13.97 & \underline{14.33} & 25.0 & 11.7 & 16.9  \\
Transformer & 13.76  & \underline{14.61}  & 14.49 & 31.4 & 13.4 & 29.8 & 12.38 & \underline{13.55} & 12.95 & 29.8 & 12.8 & 27.9  \\ 
\toprule
WMT14 & \multicolumn{6}{c|}{ De $\rightarrow$ En}  & \multicolumn{6}{c}{En $\rightarrow$ De} \\ 
\midrule
RNN  & 23.08  & \underline{24.62} & 24.54 & 38.4 & 26.6 & 36.4 & 18.87  & \underline{20.59} & 19.89 & 33.2 & 22.4 & 32.5  \\
ConvS2S  & 27.52  & \underline{28.79} & 28.56 & 22.5 & 9.9 & 14.6 & 24.86  & 25.71 & \underline{26.04} &  19.9 & 9.1 & 13.6  \\
Transformer & 26.44  & \underline{27.31}  & 26.96 & 32.9 & 14.3 & 30.9 & 22.01 & \underline{22.74} & 22.31 & 28.5 & 12.2 & 26.1 \\ 
\end{tabular}
\end{center}
\caption{Generation quality (BLEU) and speed (tokens/sec). Speed is measured for sentence-by-sentence generation without mini-batching on the test set on CPU.  We show the result by the underlying model with greedy decoding (greedy), beam search with $k=4$ (beam4) and our trainable greedy decoder (tg).}
\label{tb:bleu} 
\end{table*}

\begin{table}[t]
\centering
\small
\begin{center}
\begin{tabular}{l||cc||cc}
& \multicolumn{4}{c}{BLEU$\uparrow$}  \\ 
& \multicolumn{1}{c}{tg} & \multicolumn{1}{c}{tg+beam4} & \multicolumn{1}{c}{tg} & \multicolumn{1}{c}{tg+beam4}
\\ 
\toprule
IWSLT16 & \multicolumn{2}{c|}{ De $\rightarrow$ En}  & \multicolumn{2}{c}{En $\rightarrow$ De} \\ 
\midrule
RNN  & 23.59 & 25.03 & 19.88 & 20.72  \\
ConvS2S  & 28.74 & 29.50 & 24.42 & 24.74    \\
Transformer & 28.36 & 28.95 & 25.46 & 25.89 \\ 
\toprule
WMT15 & \multicolumn{2}{c|}{ Fi $\rightarrow$ En}  & \multicolumn{2}{c}{En $\rightarrow$ Fi} \\ 
\midrule
RNN  & 13.02 & 13.49 & 10.57 & 11.04    \\
ConvS2S  & 17.17 & 17.51 & 14.33 & 14.87 \\
Transformer & 14.49 & 14.79 & 12.95 & 13.45  \\ 
\toprule
WMT14 & \multicolumn{2}{c|}{ De $\rightarrow$ En}  & \multicolumn{2}{c}{En $\rightarrow$ De} \\ 
\midrule
RNN  & 24.54 & 24.86 & 19.89 & 20.56  \\
ConvS2S  & 28.56 & 28.46 & 26.04 & 26.08  \\
Transformer & 26.96 & 27.21 & 22.31 & 21.92 \\ 
\end{tabular}
\end{center}
\caption{Generation quality (BLEU$\uparrow$) using the proposed trainable greedy decoder without and with beam search ($k=4$). Results without beam search (\textit{tg}) are also appeared in Table~\ref{tb:bleu}.} 
\label{tb:tgbeam}
\end{table} 

\subsection{Setting}

We evaluate our approach on IWSLT16 German-English, WMT15 Finnish-English, and WMT14 De-En translation in both directions with three strong translation model architectures.

For IWSLT16, we use \textit{tst2013} and \textit{tst2014} for validation and testing, respectively. For WMT15, we use \textit{newstest2013} and \textit{newstest2015} for validation and testing, respectively. For WMT14, we use \textit{newstest2013} and \textit{newstest2014} for validation and testing, respectively. All the data are tokenized and segmented into subword symbols using byte-pair encoding  \citep[BPE;][]{Sennrich2016NeuralMT} to restrict the size of the vocabulary. Our primary evaluations use tokenized and cased BLEU. For METEOR and TER evaluations, we use multeval\footnote{https://github.com/jhclark/multeval} with tokenized and case-insensitive scoring. All the underlying models are trained from scratch, except for ConvS2S WMT14 English-German translation, for which we use the trained model (as well as training data) provided by \citet{Gehring2017ConvolutionalST}.\footnote{https://s3.amazonaws.com/fairseq-py/models/wmt14.v2.en-de.fconv-py.tar.bz2} 

\paragraph{RNN} We use OpenNMT-py \citep{klein-EtAl:2017:ACL-2017-System-Demonstrations}\footnote{https://github.com/OpenNMT/OpenNMT-py} to implement our model. It is composed of an encoder with two-layer bidirectional RNN, and a decoder with another two-layer RNN. We refer to OpenNMT's default setting ($\textit{rnn\_size}=500$, $\textit{word\_vec\_size}=500$) and the setting in \citet{artetxe2018unsupervised} ($\textit{rnn\_size}=600$, $\textit{word\_vec\_size}=300$), and choose similar hyper-parameters: $\textit{rnn\_size}=500$, $\textit{word\_vec\_size}=300$ for IWSLT16 and $\textit{rnn\_size}=600$, $\textit{word\_vec\_size}=500$ for WMT.
We use the input-feeding decoder and global attention with the general alignment function \cite{Luong2015EffectiveAT}.
\paragraph{ConvS2S} We implement our model based on fairseq-py.\footnote{https://github.com/facebookresearch/fairseq-py} We follow the settings in \textit{fconv\_iwslt\_de\_en} and \textit{fconv\_wmt\_en\_de} for IWSLT16 and WMT, respectively.
\paragraph{Transformer} We implement our model based on the code from \citet{Gu2017NonAutoregressiveNM}.\footnote{https://github.com/salesforce/nonauto-nmt} We follow their hyperparameter settings for all experiments.

In the results below, we focus on the \textit{gate} actor and pseudo-parallel corpora constructed by choosing the sentence with the best BLEU score from the $k$-best list produced by beam search with $k=35$. Experiments motivating these choices are shown later in this section.

\subsection{Results and Analysis}

\begin{table*}[!t]
	\centering
    \small
\begin{tabular}{l | l}
src & Am Vormittag wollte auch die Arbeitsgruppe Migration und Integration ihre Beratungen fortsetzen . \\
ref & During the morning , the Migration and Integration working group also sought to continue its discussions . \\
\midrule
greedy & The morning also \underline{wanted to continue} its discussions on migration and integration . \\
beam4 & In the morning , the working group on migration and integration also \underline{wanted to continue} its discussions . \\
beam35$^\star$ & In the morning , the migration and integration working group also \underline{wanted to continue} its discussions . \\
\midrule
tg & The morning , the Migration and Integration Working Group \underline{wanted to continue} its discussions . \\
tg+beam4 & In the morning , the Migration and Integration Working Group \underline{wanted to continue} its discussions . \\
\midrule \midrule
src & Die meisten Mails werden unterwegs mehrfach von Software-Robotern gelesen . \\
ref & The majority of e-mails are read several times by software robots en route to the recipient . \\
\midrule
greedy & Most mails are \underline{read by} software robots on the go . \\
beam4 & Most mails are \underline{read by} software robots on the go . \\
beam35$^\star$ & Most e-mails are \underline{read several times by} software robots on the road . \\
\midrule
tg & Most mails are \underline{read several times by} software robots on the road . \\
tg+beam4 & Most mails are \underline{read several times by} software robots on the road . \\
\midrule \midrule
src & Ich suche schon seit einiger Zeit eine neue Wohnung für meinen Mann und mich . \\
ref & I have been looking for a new home for my husband and myself for some time now . \\
\midrule
greedy & I have been looking for a new apartment \underline{for some time for my husband and myself} . \\
beam4 & I have been looking for a new apartment \underline{for some time for my husband and myself} . \\
beam35$^\star$ & I have been looking for a new apartment \underline{for my husband and myself for some time now} . \\
\midrule
tg & I have been looking for a new apartment \underline{for my husband and myself for some time now} . \\
tg+beam4 & I have been looking for a new apartment \underline{for my husband and myself for some time now} . \\
\end{tabular}
	\caption{Translation examples from the WMT14 De-En  test set with Transformer. We show translations generated by the underlying transformer using greedy decoding, beam search with $k=4$, and beam search with $k=35$ and the oracle BLEU scorer ($\star$). We also show the translations using our trainable greedy decoder both without and with beam search. Phrases of interest are underlined. 
    \label{fig:sample}
    }
\end{table*} 

\begin{table*}[t]
\begin{center}
\small
\begin{tabular}{l|c c c c|c c c c}
 & \multicolumn{4}{c|}{IWSLT16 De-En} & \multicolumn{4}{c}{WMT14 De-En}  \\
 & \multicolumn{1}{c}{ref} & \multicolumn{1}{c}{greedy} & \multicolumn{1}{c}{k35$^\star$} & \multicolumn{1}{c|}{tg} & \multicolumn{1}{c}{ref} & \multicolumn{1}{c}{greedy} & \multicolumn{1}{c}{k35$^\star$} & \multicolumn{1}{c}{tg} \\ 
 \toprule
 Base Model & 20.4 & 65.3 & 61.5 & 64.2 & 23.5 & 65.2 & 63.8 & 65.1 \\
+Trainable Greedy Decoder & 19.1 & 70.4 & 65.3 & 75.1 & 18.9 & 76.0 & 72.6 & 82.8 \\ 
\end{tabular}
\end{center}
\caption{Word-level likelihood (\%) averaged by sentence for the IWSLT16 and WMT14 De-En test sets with Transformer.
Each row represents the model used to evaluate word-level likelihood, and each column represents a different source of translations, including the reference (ref), greedy decoding on the base model (greedy), beam search with $k=35$ on the base model and the BLEU scorer (k35$^\star$), and trainable greedy decoder (tg).}\label{tb:wprob}
\end{table*} 

The results (Table \ref{tb:bleu}) show that the use of the actor makes it practical to replace beam search with greedy decoding in most cases: We lose little or no performance, and doing so yields an increase in decoding efficiency, even accounting for the small overhead added by the actor. Among the three architectures, ConvS2S---the one with the most and largest layers---performs best. We conjecture that this gives the decoder more flexibility with which to guide decoding. In cases where model throughput is less important, our method can also be combined with beam search at test time to yield results somewhat better than either could achieve alone. Table \ref{tb:tgbeam} shows the result when combining our method with beam search. 

\paragraph{Examples} Table~\ref{fig:sample} shows a few selected translations from the WMT14 German-English test set. In manual inspection of these examples and others, we find that the actor encourages models to recover missing tokens, optimize word order, and correct prepositions.

\paragraph{Likelihood}
We also compare word-level likelihood for different decoding results assigned by the base model and the actor-augmented model. For a sentence pair $\langle \mathbf{x}, \mathbf{y} \rangle$, word-level likelihood is defined as
\begin{eqnarray}
P_w =\frac{1}{T} \sum_{t=1}^T P(y_t|y_{<t},\mathbf{x}; \bm{\theta}).
\end{eqnarray}
Table \ref{tb:wprob} shows the word-level likelihood averaged over the test set for IWSLT16 and WMT14 German to English translation with Transformer. Our trainable greedy decoder learns a much more peaked distribution and assigns a much higher probability mass to its greedy decoding result than the base model. When evaluated under the base model, the translations from trainable greedy decoding have smaller likelihood than the translations from greedy decoding using the base model for both datasets. This indicates that the trainable greedy decoder is able to find a sequence that is not highly scored by the underlying model, but that corresponds to a high value of the target metric.

\begin{figure}[t]
	\centering\includegraphics[width=0.93\linewidth]{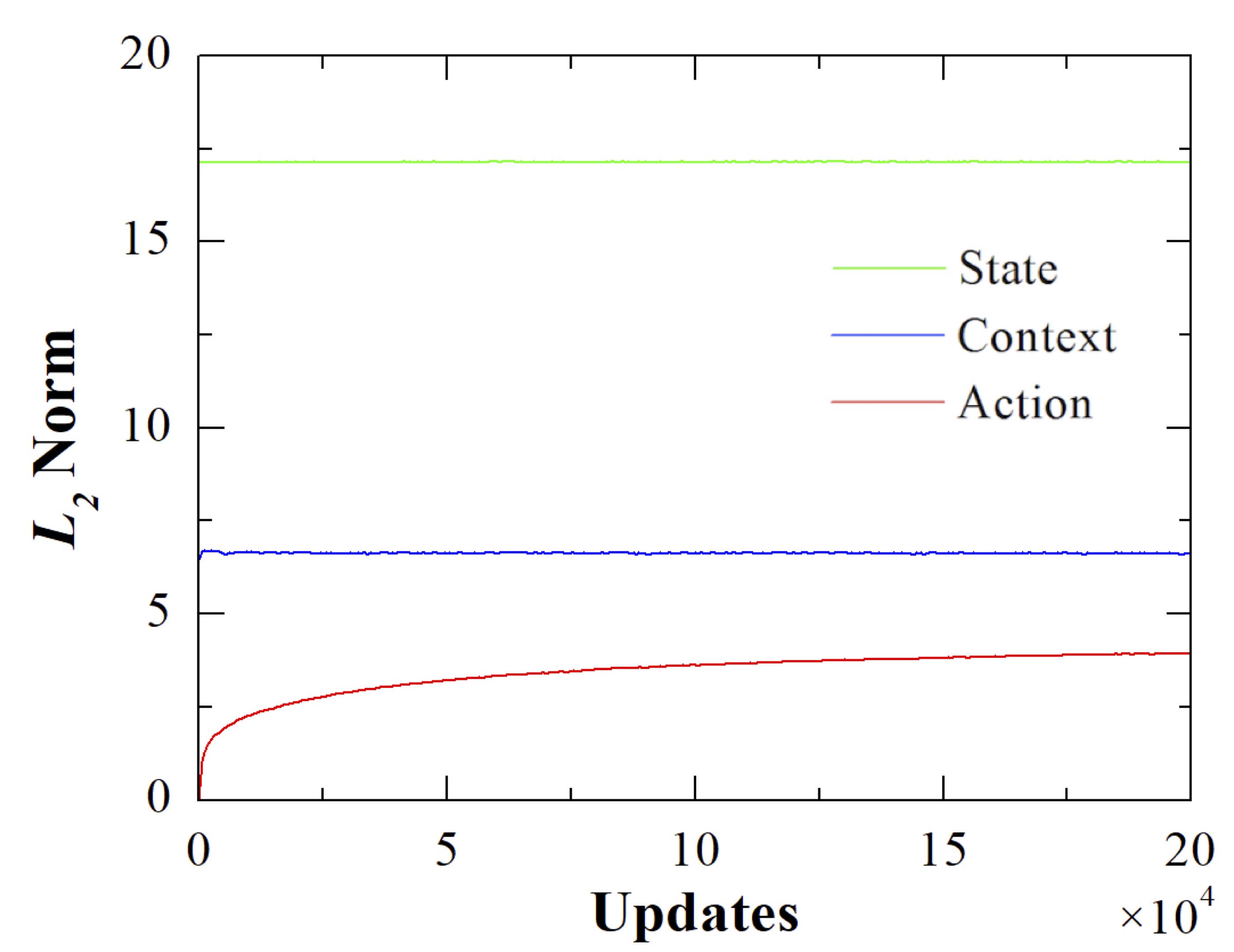}
    \caption{The norms of the three activation vectors on the IWSLT16 De-En validation set with Transformer. \textit{Action}, \textit{Context} and \textit{State} represent the norm of the action, attentional source context vector and decoder hidden state, respectively.
    }\label{fig:norm} 
\end{figure}

\paragraph{Magnitude of Action Vector} We also record the $L_2$ norm of the action, decoder hidden state, and attentional source context vectors on the validation set. Figure \ref{fig:norm} shows these values over the course of training on the IWSLT16 De-En validation set with Transformer. The norm of the action starts small, increases rapidly early in training, and converges to a value well below that of the decoder hidden state. This suggests that the action adjusts the decoder’s hidden state only slightly, rather than overwriting it.

\subsection{Effects of Model Settings}
\begin{figure}[t]
	\centering\includegraphics[width=0.9\linewidth]{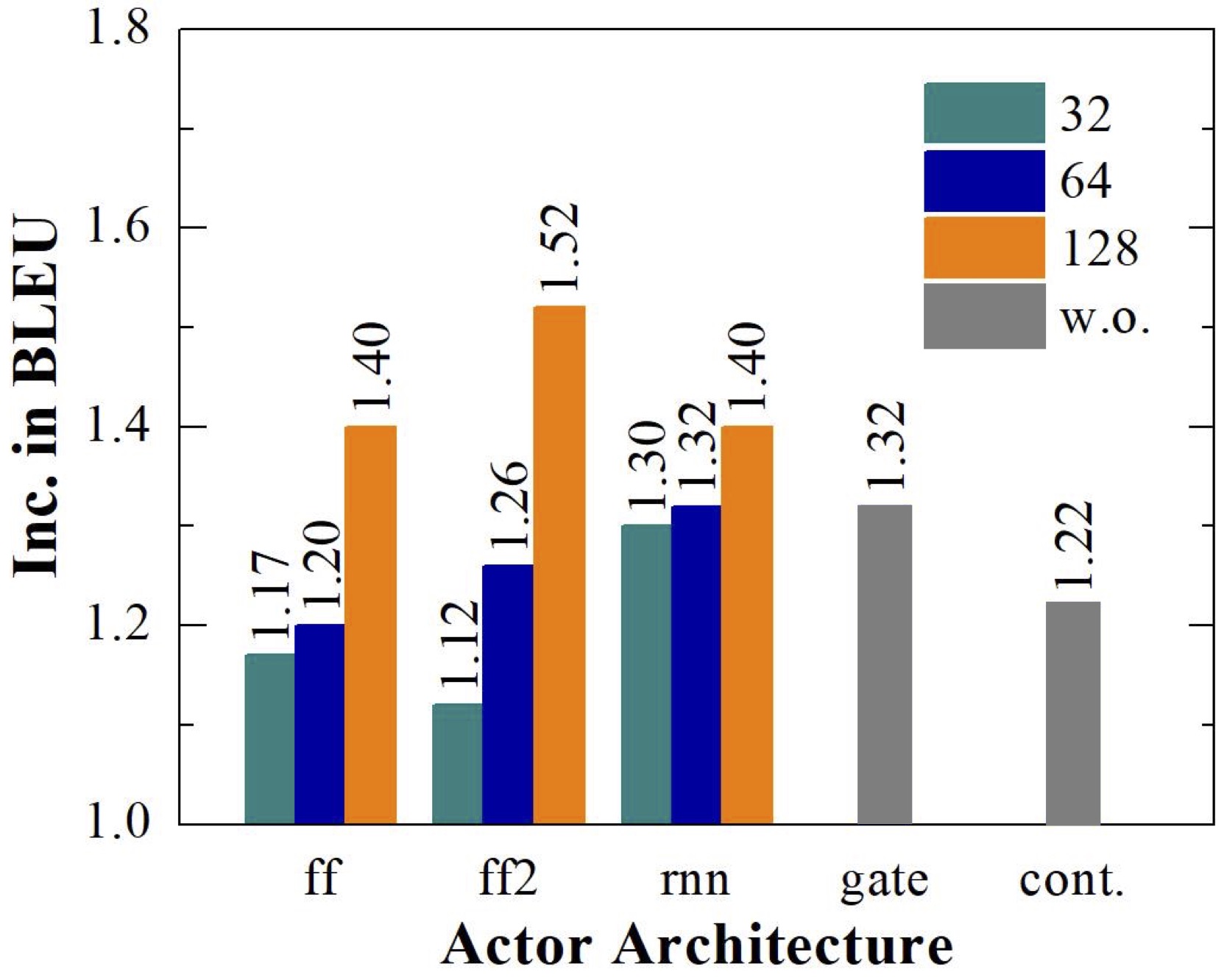}
\caption{The effect of the actor architecture and hidden state size on trainable greedy decoding results over the IWSLT16 De-En validation set with Transformer (BLEU$\uparrow$), shown with a baseline (\textit{cont.}) in which the underlying model, rather than the actor, is trained on the pseudo-parallel corpus. The Y-axis starts from 1.0. \textit{w.o.} indicates an actor with no hidden layer. 0.0 corresponds to 33.04 BLEU.}\label{fig:arch} 
\end{figure}

\paragraph{Actor Architecture}
Figure \ref{fig:arch} shows the trainable greedy decoding result on IWSLT16 De-En validation set with different actor architectures. We observe that our approach is stable across different actor architectures and is relatively insensitive to the hyperparameters of the actor.  For the same type of actor, the performance increases gradually with the hidden layer size. The use of a recurrent connection within the actor does not meaningfully improve performance, possibly since all actors can use the recurrent connections of the underlying decoder. Since the gate actor contains no additional hyperparameters and was observed to learn quickly and reliably, we use it in all other experiments. 

Here, we also explore a simple alternative to the use of the actor: creating a pseudo-parallel corpus with each model, and then training each model, unmodified and entirety, directly on this new corpus. This experiment (\textit{cont.} in Figure \ref{fig:arch}) yields results that are comparable to, but not better than, the results seen with the actors. However, this comes with substantially greater computational complexity at training time, and, if the same trained model is to be optimized for multiple target metrics, greater storage costs as well.

\paragraph{Beam Size}

\begin{figure}[t]
	\centering\includegraphics[width=0.85\linewidth]{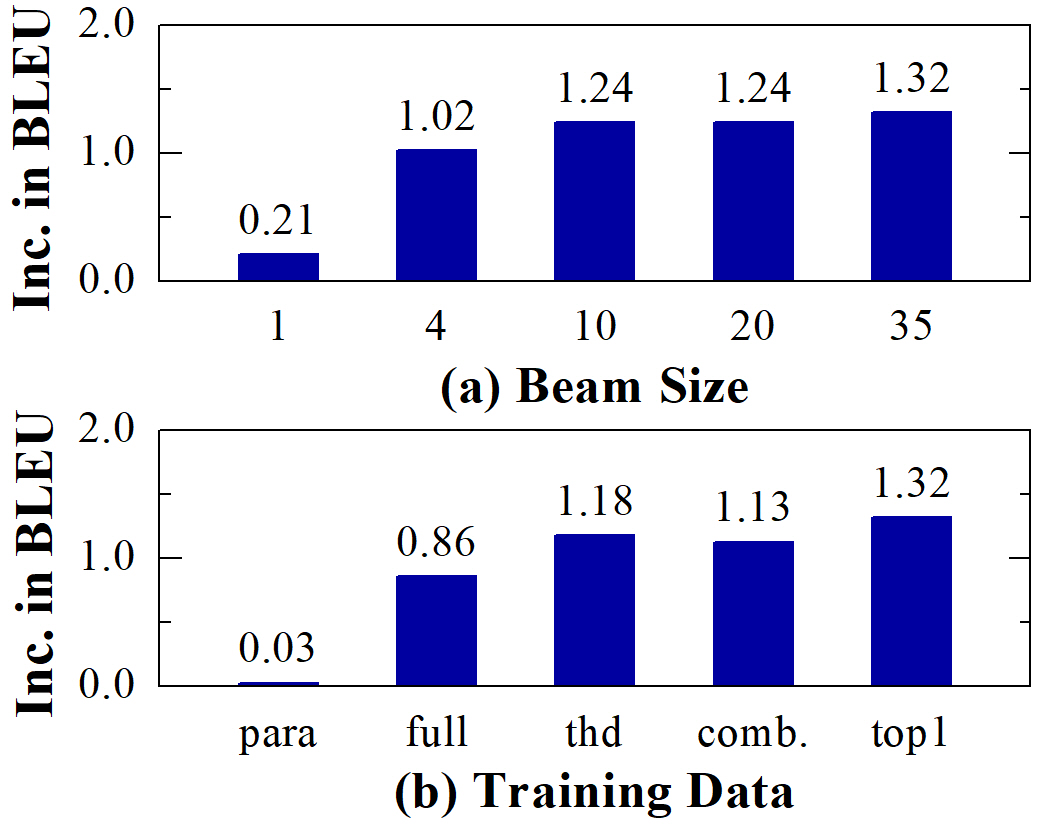} 
\caption{(a) The effect of beam size on the IWSLT16 De-En validation with Transformer and (b) the effect of the training corpus composition in the same setting. \textit{para}: parallel corpus; \textit{full}: all $35$ beam search outputs; \textit{thd}: beam search outputs that score higher than the base model's greedy decoding output; \textit{top1}: beam search output with the highest bleu score; \textit{comb}.: \textit{top1}+\textit{para}.
0.0 corresponds to 33.04 BLEU.
}\label{fig:beam}
\end{figure}

Figure~\ref{fig:beam}a shows the effect of the beam size used to generate the pseudo-parallel corpus on the IWSLT16 De-En validation set with Transformer. Trainable greedy decoding improves over greedy decoding even when we set $k=1$, namely, running \textit{greedy} decoding on the unaugmented model to construct the new training corpus. With increased beam size $k$, the BLEU score consistently increases, but we observe diminishing returns beyond roughly $k=35$, and we use that value elsewhere.

\paragraph{Training Corpus Construction}

There are a variety of ways one might use the output of beam search to construct a pseudo-parallel corpus: We could use the single highest-scoring output (by BLEU, or our target metric) for each input (\textit{top1}), use all $35$ beam search outputs (\textit{full}), use all those outputs that score higher than the threshold, namely the base model's greedy decoding output (\textit{thd}), or combine the \textit{top1} results with the gold-standard translations (\textit{comb.}). We show the effect of training corpus construction in Figure \ref{fig:beam}b. \textit{para} denotes the baseline approach of training the actor with the original parallel corpus used to train the underlying NMT model. Among the four novel approaches, \textit{full} obtains the worst performance, since the beam search outputs contain translations that are far from the gold-standard translation. We choose the best performing \textit{top1} strategy.

\paragraph{Decoding Objectives}
As our approach is capable of using an arbitrary decoding objective, we investigate the effect of different objectives on BLEU, METEOR (MTR) and TER scores with Transformer for IWSLT16 De-En translation. Table \ref{table:crit} shows the final result on the test set. When trained with one objective, our model yields relatively good performance on that objective. For example, negative sentence-level TER (i.e., -sTER) leads to -3.0 TER improvement over greedy decoding and -0.5 TER improvement over beam search. However, since these objectives are all well correlated with each other, training with different objectives do not differ dramatically. 

\section{Related Work}
\paragraph{Data Distillation} Our work is directly inspired by work on knowledge distillation, which uses a similar pseudo-parallel corpus strategy, but aims at training a compact model to approximate the function learned by a larger model or an ensemble of models \cite{hinton2015distilling}. \citet{kim-rush:2016:EMNLP2016} introduce knowledge distillation in the context of NMT, and show that a smaller student network can be trained to achieve similar performance to a teacher model by learning from pseudo-corpus generated by the teacher model. \citet{Zhang2017TowardsCA} propose a new strategy to generate a pseudo-corpus, namely, fast sequence-interpolation based on the greedy output of the teacher model and the parallel corpus. \citet{Freitag2017EnsembleDF} extend knowledge distillation on an ensemble and oracle BLEU teacher model. However, all these approaches require the expensive procedure of retraining the full student network.

\begin{table}[t]
\small 
	\centering
	\begin{tabular}{l l | c c c }
    	 & Obj. & BLEU$\uparrow$ & MTR$\uparrow$ & TER$\downarrow$   \\ \toprule 
        Greedy & - &27.15 & 29.0 & 54.4 \\ 
        Beam4 & - & 28.74 & 29.9 & 51.9 \\ 
        \midrule
         & sBLEU & \underline{28.36} & \underline{29.7} & 52.0 \\ 
        & sMTR &  \underline{28.36} & 29.6 & 51.8  \\ 
        & -sTER & 28.05 & 29.6 & \underline{51.4} \\ 
	\end{tabular}
	\caption{Results when trained with different decoding objectives  on IWSLT16 De-En translation using Transformer. MTR denotes METEOR. We report greedy decoding and beam search ($k=4$) results using the original model, and results with trainable greedy decoding (lower half).
    }\label{table:crit}
\end{table}

\paragraph{Pseudo-Parallel Corpora in Statistical MT} Pseudo-parallel corpora generated from beam search have been previously used in statistical machine translation (SMT) \cite{chiang2012hope,gao-he:2013:NAACL-HLT,auli-gao:2014:P14-2,dakwale2016improving}.  \citet{gao-he:2013:NAACL-HLT} integrate a recurrent neural network language model as an additional feature into a trained phrase-based SMT system and train it by maximizing the expected BLEU on $k$-best list from the underlying model. Our work revisits a similar idea in the context trainable greedy decoding for neural MT.

\paragraph{Decoding for Multiple Objectives} Several works have proposed to incorporate different decoding objectives into training. \citet{Ranzato2015SequenceLT} and \citet{Bahdanau2016AnAA} use reinforcement learning to achieve this goal. \citet{shen-EtAl:2016:P16-1} and \citet{Norouzi2016RewardAM} train the model by defining an objective-dependent loss function. \citet{wiseman-rush:2016:EMNLP2016} propose a learning algorithm tailored for beam search. Unlike these works that optimize the entire model, \citet{Li2017LearningTD} introduce an additional network that predicts an arbitrary decoding objective given a source sentence and a prefix of translation. This prediction is used as an auxiliary score in beam search. All of these methods focus primarily on improving beam search results, rather than those with greedy decoding.

\section{Conclusion}
This paper introduces a novel method, based on an automatically-generated pseudo-parallel corpus, for training an actor-augmented decoder to optimize for greedy decoding. Experiments on three models and three datasets show that the training strategy makes it possible to substantially improve the performance of an arbitrary neural sequence decoder on any reasonable translation metric in either greedy or beam-search decoding, all with only a few trained parameters and minimal additional training time.

As our model is agnostic to both the model architecture and the target metric, we see the exploration of more diverse and ambitious model--target metric pairs as a clear avenue for future work.

\section*{Acknowledgments}

This work was partly supported by Samsung Advanced Institute of Technology (Next Generation Deep Learning: from pattern recognition to AI), Samsung Electronics (Improving Deep Learning using Latent Structure) and the Facebook Low Resource Neural Machine Translation Award. KC thanks support by eBay, TenCent, NVIDIA and CIFAR. This project has also benefited from financial support to SB by Google and Tencent Holdings.  

\bibliography{emnlp2018}
\bibliographystyle{acl_natbib_nourl}

\end{document}